\documentclass[10pt,twocolumn,letterpaper]{article}

\usepackage[pagenumbers]{cvpr} % To force page numbers, e.g. for an arXiv version

\usepackage{graphicx}
\usepackage{amsmath}
\usepackage{amssymb}

\usepackage{algorithm}
\usepackage{algorithmic}
\usepackage{bm}
\usepackage{mathtools}
\usepackage{makecell}
\usepackage{enumitem}
\usepackage{caption}
\usepackage{lipsum}
\usepackage{cuted}
\usepackage{siunitx}
\usepackage{booktabs}

\usepackage{subcaption}
\usepackage{multirow}
\newcommand{\vect}[1]{\boldsymbol{\mathbf{#1}}}

\usepackage[pagebackref,breaklinks,colorlinks]{hyperref}

\usepackage[capitalize]{cleveref}
\crefname{section}{Sec.}{Secs.}
\Crefname{section}{Section}{Sections}
\Crefname{table}{Table}{Tables}
\crefname{table}{Tab.}{Tabs.}

\def\cvprPaperID{08687} % *** Enter the CVPR Paper ID here
\def\confName{CVPR}
\def\confYear{2022}

\begin{document}

\title{RepMLPNet: Hierarchical Vision MLP with Re-parameterized Locality}

\author{Xiaohan Ding \textsuperscript{1} \thanks{This work is supported by the National Natural Science
		Foundation of China (Nos.61925107, U1936202, 62021002) and the Beijing Academy of Artificial Intelligence (BAAI). This work is done during Xiaohan Ding and Honghao Chen's internship at MEGVII Technology.} \quad Honghao Chen \textsuperscript{2} \quad Xiangyu Zhang \textsuperscript{3} \quad Jungong Han \textsuperscript{4} \quad Guiguang Ding \textsuperscript{1}\thanks{Corresponding author.} \\
	\textsuperscript{1} Beijing National Research Center for Information Science and Technology (BNRist); \\School of Software, Tsinghua University, Beijing, China \\
	\textsuperscript{2} Institute of Automation, Chinese Academy of Sciences \\
	\textsuperscript{3} MEGVII Technology \\
	\textsuperscript{4} Computer Science Department, Aberystwyth University, SY23 3FL, UK \\
	\tt\small dxh17@mails.tsinghua.edu.cn \quad chenhonghao2021@ia.ac.cn \quad zhangxiangyu@megvii.com\\
	\tt\small jungonghan77@gmail.com \quad dinggg@tsinghua.edu.cn\\
}
\maketitle

%%%%%%%%% ABSTRACT
\begin{abstract}
	Compared to convolutional layers, fully-connected (FC) layers are better at modeling the long-range dependencies but worse at capturing the local patterns, hence usually less favored for image recognition. In this paper, we propose a methodology, Locality Injection, to incorporate local priors into an FC layer via merging the trained parameters of a parallel conv kernel into the FC kernel. Locality Injection can be viewed as a novel Structural Re-parameterization method since it equivalently converts the structures via transforming the parameters. Based on that, we propose a multi-layer-perceptron (MLP) block named RepMLP Block, which uses three FC layers to extract features, and a novel architecture named RepMLPNet. The hierarchical design distinguishes RepMLPNet from the other concurrently proposed vision MLPs. As it produces feature maps of different levels, it qualifies as a backbone model for downstream tasks like semantic segmentation. Our results reveal that 1) Locality Injection is a general methodology for MLP models; 2) RepMLPNet has favorable accuracy-efficiency trade-off compared to the other MLPs; 3) RepMLPNet is the first MLP that seamlessly transfer to Cityscapes semantic segmentation. The code and models are available at \url{https://github.com/DingXiaoH/RepMLP}.
\end{abstract}

%%%%%%%%% BODY TEXT
\section{Introduction}
\label{sec:intro}

The locality of images (\ie, a pixel is more related to its neighbors than the distant pixels) makes \emph{Convolutional Neural Network (CNN)} successful in image recognition, as a conv layer only processes a local neighborhood. In this paper, we refer to this inductive bias as the \textit{local prior}.

Besides, we also desire the ability to build up long-range dependencies, which is referred to as the \emph{global capacity} in this paper. Traditional CNNs model the long-range dependencies by deep stacks of conv layers \cite{wang2018non}. However, repeating local operations may cause optimization difficulties~\cite{he2016deep,wang2018non,cao2019gcnet}. Some prior works enhance the global capacity with self-attention-based modules \cite{wang2018non,dosovitskiy2020image,vaswani2017attention}, which have no local prior. For example, due to the lack of local prior, ViT~\cite{dosovitskiy2020image} requires an enormous amount of training data ($3\times10^8$ images in JFT-300M) to converge. On the other hand, a \emph{fully-connected (FC)} layer can also directly model the dependencies between any two input points, which is as simple as flattening the feature map as a vector, linearly mapping it into another vector, and reshaping the resultant vector back into a feature map. However, this process has no locality either. Without such an important inductive bias, the concurrently proposed MLPs~\cite{tolstikhin2021mlp,touvron2021resmlp,liu2021pay,yu2021s} usually demand a huge amount of training data (\eg, JFT-300M), more training epochs (300 or 400 ImageNet \cite{deng2009imagenet} epochs) or special training techniques (\eg, a DeiT-style distillation method \cite{touvron2021training,touvron2021resmlp}) to \emph{learn the inductive bias from scratch}. Otherwise, they may not reach a comparable level of performance with traditional CNNs. 

We desire an MLP model that is \textbf{1)} friendly to small-data, \textbf{2)} trainable with ordinary training methods, and \textbf{3)} effective in \emph{visual recognition}. To this end, we make contributions in three aspects: methodology, component and architecture.

\begin{figure*}
	\begin{center}
		\includegraphics[width=\linewidth]{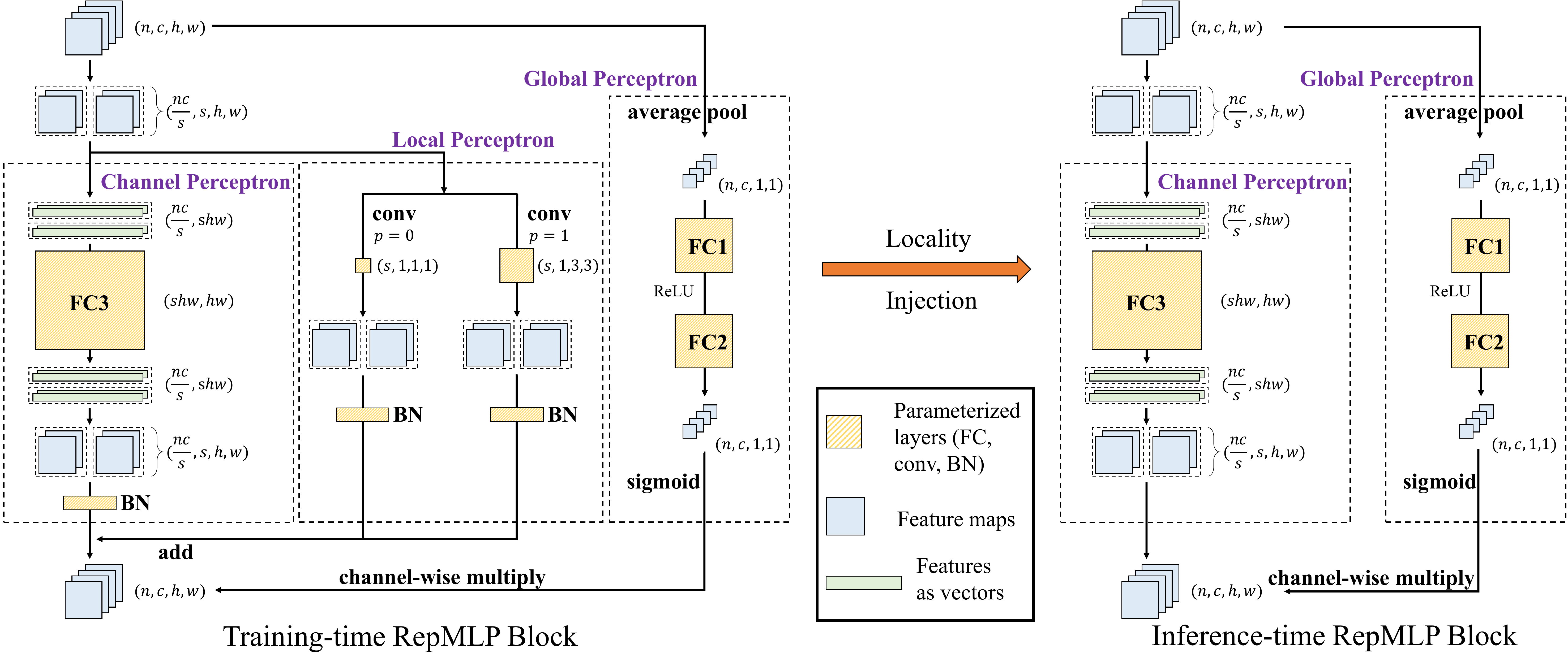}
		\vspace{-0.25in}
		\caption{RepMLP Block, where $n,c,h,w$ are the batch size, number of input channels, height and width of the feature map, $s$ is the number of \textit{share-sets}, $p$ is the padding. This example assumes $n=1$, $c=4$, $s=2$. \textbf{1)} The Global Perceptron aggregates the information across all the spatial locations and establishes the relations among channels. \textbf{2)} In parallel, the input feature map is split into $s$ share-sets and fed into the Channel Perceptron, which simply reshapes the features into vectors, linearly maps it to the output vectors, and reshapes them back. \textbf{3)} The Local Perceptron takes the same inputs as the Channel Perceptron but convolve with small kernels to capture the local patterns. This example uses 1$\times$1 and 3$\times$3 so that the padding should be $p=0$ and 1, respectively, to maintain the feature map size. Through batch normalization (BN) \cite{ioffe2015batch}, the outputs of Local Perceptron and Channel Perceptron and added up. Finally, we combine the global and channel-wise information by merging the outcomes of the Global Perceptron. After training, the conv layers are absorbed into the FC3 kernel via Locality Injection, so that the training-time block is equivalently converted into a three-FC block and used for inference.}
		\label{fig-block}
		\vspace{-0.3in}
	\end{center}
\end{figure*}

\noindent\textbf{Methodology} We propose a novel methodology, \emph{Locality Injection}, to provide an FC layer with what it demands for effective visual understanding: the locality. Specifically, we place one or more conv layers parallel to the FC and add up their outputs. Though the FC simply views the feature maps as vectors, completely ignoring the locality, the conv layers can capture the local patterns. However, though such conv layers bring only negligible parameters and compute, the inference speed may be considerably slowed down because of the reduction of degree of parallelism on high-power computing devices like GPUs~\cite{ma2018shufflenet}. Therefore, we propose to equivalently \emph{merge the conv layers into the FC kernels} after training to speed up the inference. By doing so, we obtain an FC layer that is structurally identical to a normally trained FC layer but is parameterized by a special matrix with locality. Since Locality Injection converts the training-time structure (FC + conv) to the inference-time (a single FC) via an equivalent transformation on the parameters, it can be viewed as a novel Structural Re-parameterization~\cite{ding2021repvgg,ding2019acnet,ding2021diverse,ding2021resrep,ding2022scaling} technique. In other words, we \emph{equivalently incorporate the inductive bias into a trained FC layer, instead of letting it learn from scratch}. The key to such an equivalent transformation is converting an arbitrary conv kernel to an FC kernel (\ie, a Toeplitz matrix). In this paper, we propose a simple, platform-agnostic and differentiable approach (Sec.~\ref{sect-3}).

\noindent\textbf{Component} Based on Locality Injection, we propose \emph{RepMLP Block} as an MLP building block. Fig.~\ref{fig-block} shows a training-time RepMLP Block with FC, conv and batch normalization \cite{ioffe2015batch} (BN) layers can be equivalently converted into an inference-time block with only three FC layers.

\noindent\textbf{Architecture} The hierarchical design has been shown to benefit visual understanding~\cite{simonyan2014very,he2016deep,liu2021swin}. Therefore, we propose a hierarchical MLP architecture composed of RepMLP Blocks. In other words, as the feature map size reduces, the number of channels increases. We reckon that the major obstacle for adopting hierarchical design in a ResMLP-~\cite{touvron2021resmlp} or MLP-Mixer-style~\cite{tolstikhin2021mlp} model is that the number of parameters is coupled with the feature map size~\footnote{In this paper, an MLP refers to a model that mostly uses FC layers to linearly map features from a vector to another, so that the number of parameters must be proportional to the input size and output size. By our definition, another model, CycleMLP~\cite{chen2021cyclemlp}, is not referred to as an MLP.}, so that the number of parameters of the lower-level FC layers would be several orders of magnitude greater than the higher-level layers. For example, assume the lowest-level feature maps are of 56$\times$56, an FC layer requires $56^4=9.8$M parameters to map a channel to another, without any cross-channel correlations (\ie, like depth-wise convolution). We may let all the channels share the same set of parameters, so that the layer will have a total of 9.8M parameters. However, let the highest-level feature maps be of 7$\times$7, the parameter count is only $7^4=2.4$K but the number of channels is large. Predictably, sharing so few parameters among so many channels restricts the representational capacity hence results in inferior performance. We solve this problem by a set-sharing linear mapping (Sec.~\ref{sect-block}) so that we can independently control the parameter count of each layer by letting the channels share a configurable number of parameter sets $s$. With a smaller $s$ for the lower-level layers and a larger $s$ for the higher-level ones, we can balance the model size and the representational capacity.

Another drawback of the concurrently proposed MLPs is the difficulty of transferring to the downstream tasks like semantic segmentation. For example, MLP-Mixer \cite{tolstikhin2021mlp} demonstrates satisfactory performance on ImageNet but does not qualify as the backbone of a segmentation framework like UperNet~\cite{xiao2018unified} as it aggressively embeds (\ie, downsamples) the images by 16$\times$ and repeatedly transforms the embeddings, so that it cannot produce multi-scale feature maps with different levels of semantic information. In contrast, our hierarchical design produces semantic information on different levels, so that it can be readily used as the backbone of the common downstream frameworks. 
 
In summary, with Locality Injection, RepMLP Block and a hierarchical architecture, RepMLPNet achieves favorable accuracy-efficiency trade-off with only 100 training epochs on ImageNet, compared to the other MLP models trained in 300 or 400 epochs. We also show the universality of Locality Injection as it improves the performance of not only RepMLPNet but also ResMLP~\cite{touvron2021resmlp}. Moreover, RepMLPNet shows satisfactory performance as the first attempt to transfer an MLP-style backbone to semantic segmentation.

%TODO mention some results

 %TODO fix parameter size. define MLP.

%For the application scenarios where the primary concerns are the accuracy and throughput but not the number of parameters, one may prefer FC-based models to traditional CNNs. For example, the GPU inference serves usually have tens of GBs of memory, so that the memory occupied by the parameters is minor compared to that consumed by the computations and internal feature maps.

\section{Related Work}

\subsection{From Vision Transformer to MLP}

Vision Transformers~\cite{touvron2021training,dosovitskiy2020image,liu2021swin} heavily adopt self-attention modules to capture the spatial patterns. A primary motivation for using Transformers on vision tasks is to reduce the inductive bias designed by human and let the model automatically learn a better bias from big data. Concurrent with our work, MLP-Mixer~\cite{tolstikhin2021mlp}, ResMLP~\cite{touvron2021resmlp} and gMLP~\cite{liu2021pay} are proposed. For example, MLP-Mixer alternatively mix the information across channels (channel-mixing, implemented by 1$\times$1 conv) and within channels (token-mixing). Specifically, the token-mixing component projects the feature maps along the spatial dimension (\ie, transpose the feature map tensor), feed them into a 1$\times$1 conv, and transpose the outcomes back. Therefore, the token-mixing can be viewed as an FC layer that flattens every channel as a vector, linearly maps it into another vector, and reshapes it back into a channel, completely ignoring the positional information; and all the channels share the same kernel matrix. In this way, MLP-Mixer realizes the communications among spatial locations. ResMLP~\cite{touvron2021resmlp} and gMLP~\cite{liu2021pay} present different architectures which mix the spatial information with a similar mechanism.

\subsection{Structural Re-parameterization}

The core of Locality Injection is to equivalently merge a trained conv kernel into a trained FC kernel to inject the locality, so it can be categorized into Structural Re-parameterization, which is a methodology of converting structures via transforming the parameters. A representative of Structural Re-parameterization is RepVGG~\cite{ding2021repvgg}, which is a VGG-like architecture that uses only 3$\times$3 conv and ReLU for inference. Such an inference-time architecture is equivalently converted from a training-time architecture with identity and 1$\times$1 branches. Asymmetric Convolution Block (ACB)~\cite{ding2019acnet} and Diverse Branch Block (DBB)~\cite{ding2021diverse} are two replacements for regular conv layers. Via constructing extra training-time paths (\eg, 1$\times$3, 3$\times$1, or 1$\times$1-3$\times$3), they can improve a regular CNN without extra inference costs. ResRep~\cite{ding2021resrep} uses Structural Re-parameterization for channel pruning~\cite{li2016pruning,liu2019rethinking,ding2019centripetal,ding2019approximated,ding2021manipulating} and achieves state-of-the-art results, which reduces the filters in a conv layer via constructing and pruning a following 1$\times$1 layer. RepLKNet~\cite{ding2022scaling} heavily uses very large (\eg, 31$\times$31) convolutional kernels, where Structural Re-parameterization with small kernels helps to make up the optimization issue.

Locality Injection is a remarkable attempt to generalize Re-parameterization beyond convolution. By merging a conv into an FC kernel, we bridge conv and FC with a simple, platform-agnostic and differentiable method (Sec.~\ref{sect-3}).

\section{Locality Injection via Re-parameterization}\label{sect-3}

This section derives the explicit formula (Eq.~\ref{eq-final}) to convert any given conv kernel into an FC kernel (a Toeplitz matrix), which is the key to further merging the conv into the parallel FC. The derivation can be safely skipped.

\subsection{Formulation}

In this paper, a feature map is denoted by a tensor $\mathrm{M}\in\mathbb{R}^{n\times c\times h\times w}$, where $n,c,h,w$ are the batch size, number of channels, height and width, respectively. We use $\mathrm{F}$ and $\mathrm{W}$ for the kernel of conv and FC, respectively. For the ease of re-implementation, we formulate the computation in a PyTorch-like~\cite{paszke2019pytorch} pseudo-code style. For example, the data flow through a $k\times k$ conv is formulated as
\begin{equation}
	\mathrm{M}^{(\text{out})} = \text{CONV}(\mathrm{M}^{(\text{in})}, \mathrm{F}, p) \,,
\end{equation}
where $\mathrm{M}^{(\text{out})}\in\mathbb{R}^{n\times o\times h^\prime\times w^\prime}$ is the output, $o$ is the number of output channels, $p$ is the number of pixels to pad, $\mathrm{F}\in\mathbb{R}^{o\times c\times k\times k}$ is the conv kernel (we temporarily assume the conv is dense, \ie, the number of groups is 1). From now on, we assume $h^\prime=h, w^\prime=w$ for the simplicity.

For an FC, let $p$ and $q$ be the input and output dimensions, $\mathrm{V}^{(\text{in})}\in\mathbb{R}^{n\times p}$ and $\mathrm{V}^{(\text{out})}\in\mathbb{R}^{n\times q}$ be the input and output, respectively, the kernel is $\mathrm{W}\in\mathbb{R}^{q\times p}$ and the matrix multiplication (MMUL) is formulated as 
\begin{equation}\label{eq-formulation-v}
	\mathrm{V}^{(\text{out})} = \text{MMUL}(\mathrm{V}^{(\text{in})}, \mathrm{W})=\mathrm{V}^{(\text{in})}\cdot\mathrm{W}^\intercal \,.
\end{equation}

We now focus on an FC that takes $\mathrm{M}^{(\text{in})}$ as input and outputs $\mathrm{M}^{(\text{out})}$ and assume the output shape is the same as the input. We use $\text{RS}$ (short for ``reshape'') as the function that only changes the shape specification of tensors but not the order of data in memory, which is \textit{cost-free}. The input is first flattened into $n$ vectors of length $chw$, which is $\mathrm{V}^{(\text{in})}=\text{RS}(\mathrm{M}^{(\text{in})}, (n,chw))$, multiplied by the kernel $\mathrm{W}(ohw, chw)$, then the output $\mathrm{V}^{(\text{out})}(n, ohw)$ is reshaped back into $\mathrm{M}^{(\text{out})}(n,o,h,w)$. For the better readability, we omit the RS if there is no ambiguity, 
\begin{equation}
	\mathrm{M}^{(\text{out})}=\text{MMUL}(\mathrm{M}^{(\text{in})},\mathrm{W}) \,.
\end{equation}
Obviously, such an FC cannot take advantage of the locality of images as it computes each output point according to every input point, unaware of the positional information.

\subsection{Locality Injection}

Assume there is a conv layer parallel to the FC (like the Channel Perceptron and Local Perceptron shown in Fig.~\ref{fig-block}), which takes $\mathrm{M}^{(\text{in})}$ as input, we describe how to equivalently merge it into the FC. In the following, we assume the FC kernel is $\mathrm{W}(ohw,chw)$, conv kernel is $\mathrm{F}(o,c,k,k)$, padding is $p$. Formally, we desire to construct $\mathrm{W}^\prime$ so that
\begin{equation}
	\begin{aligned}
		&\text{MMUL}(\mathrm{M}^{(\text{in})},\mathrm{W}^\prime) \\
		&=\text{MMUL}(\mathrm{M}^{(\text{in})},\mathrm{W}) + \text{CONV}(\mathrm{M}^{(\text{in})},\mathrm{F},p) \,.
	\end{aligned}
\end{equation}

We note that for any kernel $\mathrm{W}^{(2)}$ of the same shape as $\mathrm{W}^{(1)}$, the additivity of MMUL ensures that
\begin{equation}
	\begin{aligned}
		&\text{MMUL}(\mathrm{M}^{(\text{in})}, \mathrm{W}^{(1)}) + \text{MMUL}(\mathrm{M}^{(\text{in})}, \mathrm{W}^{(2)}) \\
		&= \text{MMUL}(\mathrm{M}^{(\text{in})}, \mathrm{W}^{(1)} + \mathrm{W}^{(2)}) \,.
	\end{aligned}
\end{equation}

Therefore, we can merge $\mathrm{F}$ into $\mathrm{W}$ if we can construct $\mathrm{W}^{(\mathrm{F},p)}$ of the same shape as $\mathrm{W}$ which satisfies
\begin{equation}\label{eq-equivalency}
	\text{MMUL}(\mathrm{M}^{(\text{in})}, \mathrm{W}^{(\mathrm{F},p)}) = \text{CONV}(\mathrm{M}^{(\text{in})},\mathrm{F},p) \,.
\end{equation}
Obviously, for \textit{any} $\mathrm{M}^{(\text{in})},\mathrm{F}, p$, the corresponding $\mathrm{W}^{(\mathrm{F},p)}$ \emph{must exist}, since a conv can be viewed as a sparse FC that shares parameters among spatial positions (\ie, a Toeplitz matrix), but it is nontrivial to construct it with a given $\mathrm{F}$.

Then we seek for the explicit formula of $\mathrm{W}^{(\mathrm{F},p)}$. With the formulation used before (Eq. \ref{eq-formulation-v}), we have
\begin{equation}\label{eq-middle}
	\mathrm{V}^{(\text{out})} = \mathrm{V}^{(\text{in})}\cdot\mathrm{W}^{(\mathrm{F},p)\intercal} \,.
\end{equation}
We insert an identity matrix $\mathrm{I}$ $(chw, chw)$ and use the associative law
\begin{equation}\label{eq-insert-I}
	\mathrm{V}^{(\text{out})} = \mathrm{V}^{(\text{in})}\cdot (\mathrm{I} \cdot \mathrm{W}^{(\mathrm{F},p)\intercal}) \,.
\end{equation}
With explicit RS, we rewrite Eq. \ref{eq-insert-I} as
\begin{equation}\label{eq-last}
	\mathrm{V}^{(\text{out})} = \mathrm{V}^{(\text{in})} \cdot \text{RS}(\mathrm{I}\cdot\mathrm{W}^{(\mathrm{F},p)\intercal}, (chw, ohw)) \,.
\end{equation}

We note that $\mathrm{W}^{(\mathrm{F},p)}$ is constructed with an existing conv kernel $\mathrm{F}$, so that $\mathrm{I} \cdot \mathrm{W}^{(\mathrm{F},p)\intercal}$ \textit{is exactly a convolution} with $\mathrm{F}$ on a feature map $\mathrm{M}^{(\mathrm{I})}$ which is reshaped from $\mathrm{I}$. That is
\begin{equation}\label{eq-second-last}
	\mathrm{I}\cdot\mathrm{W}^{(\mathrm{F},p)\intercal} = \text{CONV}(\mathrm{M}^{(\mathrm{I})}, \mathrm{F}, p) \,,
\end{equation}
where $\mathrm{M}^{(\mathrm{I})}$ is reshaped from a constructed identity matrix
\begin{equation}
	\mathrm{M}^{(\mathrm{I})} = \text{RS}(\mathrm{I}, (chw, c, h, w)) \,.
\end{equation}

Comparing Eq. \ref{eq-middle} with Eq.~\ref{eq-last}, Eq.~\ref{eq-second-last}, we have
\begin{equation}\label{eq-final}
	\mathrm{W}^{(\mathrm{F},p)} = \text{RS}(\text{CONV}(\mathrm{M}^{(\mathrm{I})}, \mathrm{F}, p), (chw, ohw))^\intercal \,,
\end{equation}
which is exactly the formula to construct $\mathrm{W}^{(\mathrm{F},p)}$ with $\mathrm{F}, p$. In brief, the equivalent FC kernel of a conv kernel is the result of \emph{convolution on an identity matrix with proper reshaping}. Better still, the conversion is efficient and \textit{differentiable}, so one may derive the FC kernel during training and use it in the objective function (\eg, for penalty-based pruning \cite{han2015learning,ding2019global}). The formulas for the group-wise case are derived similarly and the code is released on GitHub.

\noindent\textbf{Why is this solution nontrivial?} Given $\mathrm{F}, p$, there is no existing generic method to construct the corresponding Toeplitz matrix $\mathrm{W}^{(\mathrm{F},p)}$ as the existing conv implementations do not require such a step. As discussed above, though $\mathrm{W}^{(\mathrm{F},p)}$ must exist, it is nontrivial to construct it in a platform-agnostic way. Modern platforms use different algorithms of conv, \eg, based on iGEMM~\cite{ding2022scaling} (for very large kernel), Winograd~\cite{winograd} (for 3$\times$3), im2col~\cite{im2col} (converting the feature map, rather than the kernel, to a matrix), FFT~\cite{fft-conv}, MEC~\cite{cho2017mec} and sliding-window). Moreover, the memory allocation of data and implementations of padding may be different. Therefore, given $\mathrm{F}, p$ and the FC kernel $\mathrm{W}^{(\mathrm{F},p)}$, the equivalency (Eq.~\ref{eq-equivalency}) may hold on a platform but break on another platform (\eg, the simplest case is the two platforms arrange the kernels in memory differently). Our method (Eq.~\ref{eq-final}) is platform-agnostic because its derivation does not rely on the concrete form of CONV. On any platform, the $\mathrm{W}^{(\mathrm{F},p)}$ constructed with its specific CONV implementation must ensure Eq.~\ref{eq-equivalency} with the same CONV.

\section{RepMLPNet}

\begin{figure}
	\begin{center}
		\includegraphics[width=\linewidth]{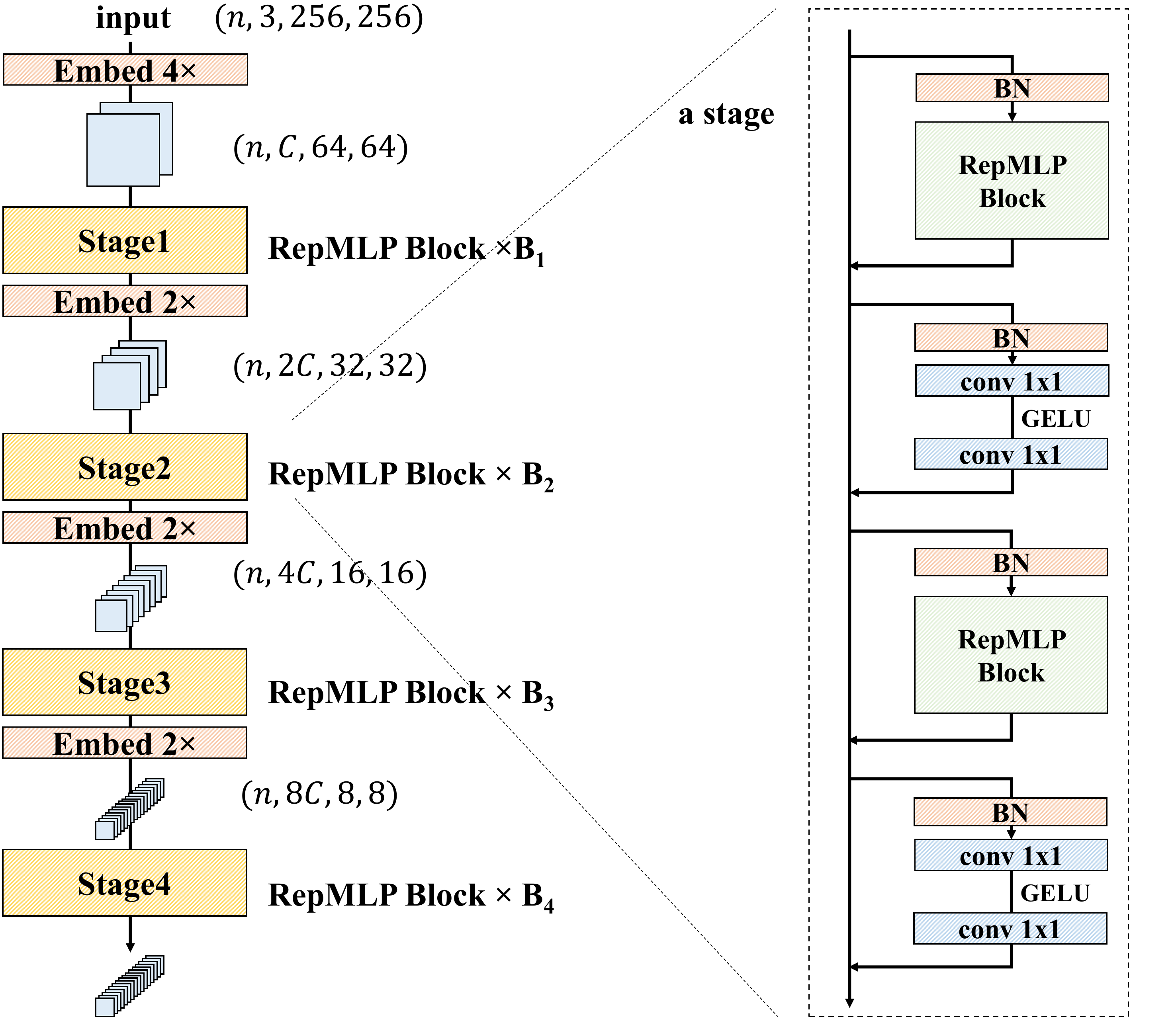}
		\vspace{-0.25in}
		\caption{Architecture of RepMLPNet. Apart from RepMLP Blocks, we also use a FFN-style block (1$\times$1-GELU~\cite{hendrycks2016gaussian}-1$\times$1) to increase the depth, which has been widely used in Vision Transformers~\cite{liu2021swin,dosovitskiy2020image}, MLPs~\cite{tolstikhin2021mlp,touvron2021resmlp} and CNN (RepLKNet~\cite{ding2022scaling}).}
		\label{fig-net}
		\vspace{-0.4in}
	\end{center}
\end{figure}

RepMLPNet is a hierarchical MLP-style architecture composed of RepMLP Blocks. We first introduce RepMLP Block (Fig. \ref{fig-block}) in Sec.~\ref{sect-block} and then describe the overall architecture (Fig. \ref{fig-net}) in  Sec.~\ref{sect-net}.

\subsection{Components of RepMLP Block}\label{sect-block}

A training-time RepMLP Block is composed of three parts termed as Global Perceptron, Channel Perceptron and Local Perceptron (Fig.~\ref{fig-block}), which are designed to model the information on different levels. Global Perceptron models the coarse global dependencies across spatial locations among all the channels; Channel Perceptron is designed for modeling the long-range spatial dependencies within each channel; Local Perceptron captures the local patterns. The outputs of the three components are combined to obtain a comprehensive transformation of the input features.

\noindent\textbf{Global Perceptron} average-pools the inputs $(n,c,h,w)$ into vectors $(n,c,1,1)$ and feed them into two FC layers to obtain a vector that encodes the global information.

\noindent\textbf{Channel Perceptron} contains an FC layer that directly performs the feature transformation, where the key is the \textit{set-sharing} mechanism. We follow the formulation in Sec.~\ref{sect-3} and assume $o=c$ for the convenience. We note that a normal FC layer has $(chw)^2$ parameters. Taking a 64$\times$64 feature map with 128 channels for example, the parameter count of a normal dense FC will be 2.1B, which is unacceptable. A natural solution is to make the FC ``depth-wise'' just like depth-wise conv, which will not be able to model cross-channel dependencies but has only $1/c$ parameters and FLOPs. However, even a parameter count of $c(hw)^2$ is too large. Our solution is to make multiple channels share a set of spatial-mapping parameters, so that the parameters are reduced to $s(hw)^2$, where $s$ is the number of \textit{share-sets} of parameters. In other words, every $\frac{c}{s}$ channels share the same set of parameters, and there are $s$ such share-sets in total. Specifically, we first evenly split the $c$ channels into $c/s$ groups, which means $(n,c,h,w) \to (\frac{nc}{s}, s, h, w)$, and then flatten them into $\frac{nc}{s}$ vectors each of length $shw$, feed the vectors into a ``depth-wise'' FC, and reshape the output back. Compared to ``depth-wise'' FC, set-sharing FC not only breaks the correlation between channels ($(chw)^2$ parameters $\to$ $c(hw)^2)$), but reduces the parameters even further ($c(hw)^2)$ $\to$ $s(hw)^2)$; but it does not reduce the computations compared to a ``depth-wise'' FC. It should be noted that the spatial mappings in ResMLP and MLP-Mixer are implemented in a different way (transpose, 1$\times$1 conv and transpose back) but are mathematically equivalent to a special case of set-sharing FC with $s=1$, which means all the channels share the same $(hw)^2$ parameters. We will show increasing $s$ can improve the performance with more parameters but no extra FLOPs, which is useful in scenarios where the model size is not a major concern (Table~\ref{table-ablation-studies}). 

In practice, though set-sharing FC is not directly supported by some computing frameworks like PyTorch, it can be implemented by a group-wise 1$\times$1 conv. Formally, let $\mathrm{V}^{(\text{in})} (\frac{nc}{s}, shw)$ be the vectors split in share-sets, the implementation is composed of three steps: \textbf{1)} reshaping $\mathrm{V}^{(\text{in})}$ as a ``feature map'' with spatial size of 1$\times$1, which is $(\frac{nc}{s}, shw, 1, 1)$; \textbf{2)} performing 1$\times$1 conv with $s$ groups (so that the parameters are $(shw)^2 / s=s(hw)^2$); \textbf{3)} reshaping the output into $(\frac{nc}{s}, s, h, w)$, then $(n, c, h, w)$.

\noindent\textbf{Local Perceptron} takes the same inputs as Channel Perceptron. Each conv (with a following BN~\cite{ioffe2015batch} as a common practice) is depth-wise and has the same number of share-sets $s$ on the $s$-channel inputs, so the kernel is $(s, 1, k, k)$. 

\noindent\textbf{Merging} the Local Perceptron into Channel Perceptron via Locality Injection requires fusing the BN into the preceding conv layers or FC3. Note the conv layers are depth-wise and the number of channels is $s$. Let $\mathrm{F}\in\mathbb{R}^{s\times1\times k\times k}$ be the conv kernel, $\vect{\mu},\vect{\sigma},\vect{\gamma},\vect{\beta}\in\mathbb{R}^{s}$ be the accumulated mean, standard deviation and learned scaling factor and bias of the following BN, we construct the kernel $\mathrm{F}^\prime$ and bias $\mathbf{b}^\prime$ as
\begin{equation}\label{eq-fuse-bn}
	\mathrm{F}^\prime_{i,:,:,:} = \frac{\vect{\gamma}_i}{\vect{\sigma}_i}\mathrm{F}_{i,:,:,:} \,,\quad \mathbf{b}^\prime_i = -\frac{\vect{\mu}_i \vect{\gamma}_i}{\vect{\sigma}_i} + \vect{\beta}_i \,.
\end{equation}
Then it is easy to verify the equivalence:
\begin{equation}
	\begin{aligned}
		&\frac{\vect{\gamma}_i}{\vect{\sigma}_i}(\text{CONV}(\mathrm{M}, \mathrm{F}, p)_{:,i,:,:} - \vect{\mu}_i) + \vect{\beta}_i \\
		&= \text{CONV}(\mathrm{M}, \mathrm{F}^\prime, p)_{:,i,:,:} + \mathbf{b}^\prime_i \,, \forall 1\leq i \leq s \,,
	\end{aligned}
\end{equation}
where the left side is the original computation flow of a conv-BN, and the right is the constructed conv with bias.

The FC3 and BN in Channel Perceptron are fused in a similar way into $\hat{\mathrm{W}}\in\mathbb{R}^{shw\times hw}$, $\hat{\mathbf{b}}\in\mathbb{R}^{shw}$. Then we convert every conv via Eq.~\ref{eq-final} and add the resultant matrix onto $\hat{\mathrm{W}}$. The biases of conv are simply replicated by $hw$ times (because all the points on the same channel share a bias value) and added onto $\hat{\mathbf{b}}$. Finally, we obtain a single FC kernel and a single bias vector, which will be used to parameterize the inference-time FC3.

\subsection{Hierarchical Architectural Design}\label{sect-net}

Some recent vision MLP models~\cite{touvron2021resmlp,tolstikhin2021mlp} show a similar design pattern: downsampling the inputs aggressively (\eg, by 16$\times$) at the very beginning, and stacking multiple blocks to process the small-sized features. In contrast, we adopt a hierarchical design, which has proven effective by previous studies on CNNs~\cite{he2016deep,xie2017aggregated,efficientnet,regnet} and Transformers~\cite{liu2021swin}.

Specifically, we arrange RepMLP Blocks in four stages, and the blocks in a stage share the same structural hyper-parameters. The input images are downsampled by 4$\times$ with an embedding layer, which is implemented by a conv layer with a kernel size of 4$\times$4 and stride of 4. From a stage to the next, we use an embedding layer to halve the width and height of feature maps and double the channels. Therefore, a RepMLPNet can be instantiated with the following hyper-parameters: the number of RepMLP Blocks in each stage $[B_1,B_2,B_3,B_4]$, the number of channels of the first stage $C$ (the four stages will have $C$, $2C$, $4C$, $8C$ channels, respectively), the input resolution $H$$\times$$W$ (so that $h_1=H/4, w_1=W/4, ..., h_4=H/32, w_4=W/32$), and the number of share-sets of each stage ($[S_1,S_2,S_3,S_4]$). Considering the number of parameters in FC3 is $s(hw)^2$, we use a smaller $s$ at an earlier stage. 

An advantage of our hierarchical architecture is that the feature maps produced by any stage can be readily used by a downstream framework. For example, UperNet \cite{xiao2018unified} requires four levels of feature maps with different sizes, so it cannot use MLP-Mixer or ResMLP as backbone.

\section{Experiments}

%In this section, we 1) present a series of RepMLPNets and compare with the other MLP models on ImageNet; 2) validate that Locality Injection is a generic methodology that works on other MLP models; 3) show that RepMLPNet can be transferred to semantic segmentation.

\subsection{ImageNet Classification}

\setlength{\tabcolsep}{4pt}
\begin{table}
	\caption{Architectural hyper-parameters of RepMLPNet models (T for tiny, B for base, D for deep, L for large). Of note is that RepMLP-D256 is deeper than RepMLP-B256 but narrower so that they have comparable FLOPs and number of parameters.}
	\label{table-model-arch}
	\vspace{-0.25in}
	\begin{center}
		\small
		\begin{tabular}{lccccc}
			\hline
			Name			&	Input		&	$\vect{B}$		&	 $C$ 		&	$\vect{S}$		\\
			\hline
			RepMLP-T224		&	224$\times$224			&	$[2,2,6,2]$		&	64		&	$[1,4,16,128]$		\\
			RepMLP-B224		&	224$\times$224			&	$[2,2,12,2]$	&	96		&	$[1,4,32,128]$		\\
			\hline
			RepMLP-T256		&	256$\times$256			&	$[2,2,6,2]$		&	64		&	$[1,4,16,128]$		\\
			RepMLP-B256		&	256$\times$256			&	$[2,2,12,2]$	&	96		&	$[1,4,32,128]$		\\
			RepMLP-D256		&	256$\times$256			&	$[2,2,18,2]$	&	80		&	$[1,4,16,128]$		\\
			RepMLP-L256		&	256$\times$256			&	$[2,2,18,2]$	&	96		&	$[1,4,32,256]$		\\
			\hline
		\end{tabular}
	\end{center}
	\vspace{-0.3in}
\end{table}
\setlength{\tabcolsep}{1.4pt}

\setlength{\tabcolsep}{4pt}
\begin{table*}
	\caption{ImageNet Results. The throughput (samples/second) is tested on the same 2080Ti GPU with a batch size of 128.}
	\label{table-imagenet}
	\vspace{-0.25in}
	\begin{center}
		\small
		\begin{tabular}{lccccccccc}
			\hline
			Model			&	Input resolution	&	Train epochs	&	Top-1 acc	&	FLOPs (B)	&	Params (M)	&	Throughput\\
			\hline
			\textbf{RepMLPNet-T224}		&	224		&	\textbf{100}	&	76.4	&	2.8		&	38.3		&	1709	\\
			ResMLP-S12 (our impl)		&	224		&	120				&	70.4			&	3.0			&	15.4		&	1895	\\		
			ResMLP-S12	\cite{touvron2021resmlp}				&	224		&	120				&	67.7			&	3.0			&	15.4		&	-\\
			ResMLP-S12 + DeiT-train\cite{touvron2021resmlp}		&	224		&	400				&	76.6			&	3.0			&	15.4		&	-\\
			\textbf{RepMLPNet-T256}		&	256		&	\textbf{100}	&	\textbf{77.5}	&	4.2		&	58.7		&	1374	\\
			\hline
			ResMLP-S24 + DeiT-train\cite{touvron2021resmlp}		&	224		&	400				&	79.4			&	6.0			&	30.0		&	961		\\
			RegNetX-6.4GF\cite{regnet}				&	224		&	120				&	79.6		&	6.4			&	26.2		&	589		\\
			\textbf{RepMLPNet-B224}		&	224	&	\textbf{100}	&	\textbf{80.1}	&	6.7		&	68.2		&	816		\\
			\hline
			ResNeXt-101	\cite{xie2017aggregated}				&	224		&	120				&	80.2		&	8.0			&	44.1		&	450		\\
			ResNet-101	\cite{he2016deep}				&	224		&	120				&	79.4		&	8.1			&	44.4		&	606		\\
			\textbf{RepMLPNet-D256}		&	256		&	\textbf{100}	&	80.8	&	8.6		&	86.9		&	715	\\
			\textbf{RepMLPNet-B256}		&	256		&	\textbf{100}	&	\textbf{81.0}	&	9.6		&	96.5		&	708	\\
			S$^2$-MLP-deep\cite{yu2021s}				&	224		&	300				&	80.7			&	10.5		&	51			&	-\\
			\hline
			\textbf{RepMLPNet-L256}		&	256		&	\textbf{100}	&	\textbf{81.7}			&	11.5		&	117.7		&	588	\\
			MLP-Mixer-B/16\cite{tolstikhin2021mlp}				&	224		&	300				&	76.4			&	12.6		&	59			&	-\\
			S$^2$-MLP-wide\cite{yu2021s}				&	224		&	300				&	80.0			&	14.0		&	71			&	-\\
			MLP-Mixer-B/16 (our impl)	&	224		&	\textbf{100}	&	76.7			&	12.6		&	59.8		&	632	\\
			gMLP-B\cite{liu2021pay}						&	224		&	300				&	81.6			&	15.8		&	73			&	- \\
			MLP-Mixer-B/16 (our impl)	&	256		&	\textbf{100}	&	77.0			&	16.4		&	60.4		&	578	\\
			ResMLP-B24 + DeiT-train	\cite{touvron2021resmlp}	&	224		&	400				&	81.0			&	23.0		&	115.7		&	-\\
			\hline
		\end{tabular}
	\end{center}
	\vspace{-0.3in}
\end{table*}
\setlength{\tabcolsep}{1.4pt}

\setlength{\tabcolsep}{4pt}
\begin{table}
	\caption{Comparisons with EfficientNets. The throughput (samples/second) is tested on 2080Ti GPU with a batch size of 128.}
	\label{table-vs-efficientnet}
	\vspace{-0.25in}
	\begin{center}
		\small
		\begin{tabular}{lccccccccc}
			\hline
			Model			&	Input	&	Top-1 acc	&	FLOPs (B)	&	Throughput\\
			\hline
			\textbf{RepMLPNet-T224}		&	224		&	\textbf{76.4}		&	2.8		&	\textbf{1709}	\\
			EfficientNet-B1				&	240		&	76.3		&	\textbf{0.7}		&	912		\\
			\hline
			\textbf{RepMLPNet-B256}		&	256		&	\textbf{81.0}		&	9.6		&	\textbf{708}		\\
			EfficientNet-B2				&	260		&	77.4		&	\textbf{1.0}		&	707		\\
			\hline
		\end{tabular}
	\end{center}
	\vspace{-0.2in}
\end{table}
\setlength{\tabcolsep}{1.4pt}

We first instantiate a series of RepMLPNets with different architectural hyper-parameters, as shown in Table~\ref{table-model-arch}.

We evaluate RepMLPNets on ImageNet~\cite{deng2009imagenet}. All the RepMLPNets are trained with identical settings: a global batch size of 256 distributed on 8 GPUs, AdamW \cite{loshchilov2017decoupled} optimizer with initial learning rate of 0.002, weight decay of 0.1 and momentum of 0.9. We train for only \textit{100 epochs} in total with cosine learning rate annealing, including a 10-epoch warm-up at the beginning. We use label smoothing of 0.1, mixup \cite{zhang2017mixup} with $\alpha=0.4$, CutMix \cite{yun2019cutmix} with $\alpha=1.0$, and RandAugment \cite{cubuk2020randaugment}. As a series of strong baselines, we present ResNet-101~\cite{he2016deep}, ResNeXt-101~\cite{xie2017aggregated} and RegNetX-6.4GF~\cite{regnet} trained with a \textit{strong scheme} with RandAugment, mixup and label smoothing. We would like the comparison to be slightly biased towards the traditional CNNs, so we train them for 120 epochs. All the models are evaluated with single crop and the throughput (samples/second) is tested on 2080Ti GPU with a batch size of 128. The BN layers in CNNs are also fused for the fair comparison.

It should be noted that most of the results reported by \cite{touvron2021resmlp,tolstikhin2021mlp,yu2021s,liu2021pay} are produced with a long training schedule of 300 or 400 epochs, or an advanced distillation method (the DeiT-style training \cite{touvron2021training}). Therefore, except for the results cited from the corresponding papers, we train an MLP-Mixer and a ResMLP-S12 with simple training settings for a fair comparison (labeled as ``our impl'' in Table~\ref{table-imagenet}). Specifically, the MLP-Mixer is trained with the settings identical to RepMLPNets; the ResMLP-S12 is trained with the official code and the same hyper-parameters as its reported 120-epoch result~\cite{touvron2021resmlp} except that we use a smaller batch size due to limited resources (but our reproduced accuracy is 2.7\% higher than its reported result~\cite{touvron2021resmlp}). 

Compared to the other CNNs and MLPs, we make the following observations. \textbf{1)} With fair settings, RepMLPNet shows superiority over the other MLPs, \eg, RepMLPNet-T256 outperforms MLP-Mixer (our impl, 256$\times$256 inputs) by 0.5\% in the accuracy while the FLOPs of the former is only 1/4 of the latter. \textbf{2)} With simple training methods, ResMLP and MLP-Mixer significantly degrade, \eg, the accuracy of ResMLP-S12 drops by 8.9\% (76.6\% $\to$ 67.7\%) without the 300-epoch DeiT-style training. \textbf{3)} RepMLPNet with 100-epoch training delivers a favorable accuracy-efficiency trade-off: RepMLPNet-B256 matches the accuracy of ResMLP-B24 without DeiT-style distillation, consumes 1/4 training epochs, has only 40\% FLOPs and fewer parameters. \textbf{4)} With the comparable FLOPs, MLPs are faster than CNNs, \eg, RepMLPNet-D256 has higher FLOPs than ResNeXt-101 but runs 1.6$\times$ as fast as the latter. This discovery suggests that MLPs are promising as high-throughput inference models.

To further demonstrate that FLOPs may not reflect the throughput~\cite{ding2021repvgg}, we train EfficientNet-B1/B2~\cite{efficientnet} with the aforementioned 120-epoch strong scheme and report the results in Table~\ref{table-vs-efficientnet}. Interestingly, though EfficientNets have very low FLOPs, the actual performance on GPU is inferior: RepMLPNet-T224 has 4$\times$ FLOPs as EfficientNet-B1 but runs 1.9$\times$ as fast as the latter; with comparable throughput, the accuracy of RepMLPNet-B256 is 3.6\% higher than EfficientNet-B2. We reckon the high throughput of RepMLPNet on GPU can be attributed to not only the efficiency of matrix multiplication but also the simplicity of architecture hence high degree of parallelism~\cite{ma2018shufflenet}.

We then study the effects of two key designs in RepMLP Block: the Global Perceptron and the set-sharing linear mapping of FC3. We increase the number of share-sets $\vect{S}$ or ablate the Global Perceptron and observe the performance as well as the model size. Fig.~\ref{table-ablation-studies} shows that Global Perceptron only adds negligible parameters and FLOPs (0.5M) but improves the accuracy by around 1\%. This is expected as the Local Perceptron and Channel Perceptron do not communicate information across channels, which is compensated by Global Perceptron. By increasing $\vect{S}$, fewer channels will be sharing the same set of mapping parameters, resulting in a significant performance gain without any extra FLOPs. Therefore, for the application scenarios where the speed-accuracy trade-off is the primary concern while the model size is not (\eg, in high-power computing centers), we may increase $\vect{S}$ for higher accuracy.

\setlength{\tabcolsep}{4pt}
\begin{table}
	\caption{Ablation studies on RepMLP-T224. }
	\label{table-ablation-studies}
	\vspace{-0.25in}
	\begin{center}
		\small
		\begin{tabular}{cccccccccc}
			\hline
				$\vect{S}$	&	Global Perceptron	&	Top-1 acc	&	FLOPs	&	Params\\
			\hline
				$[1,4,16,128$]	&					&	75.78		&	2.7B		&	37.8M\\
				$[1,4,16,128$]	&	\checkmark		&	76.48		&	+0.5M		&	38.3M\\
				$[2,8,32,256$]	&					&	75.94		&	2.7B		&	66.7M\\
				$[2,8,32,256$]	&	\checkmark		&	77.19		&	+0.5M		&	67.2M\\		
			\hline
		\end{tabular}
	\end{center}
	\vspace{-0.35in}
\end{table}
\setlength{\tabcolsep}{1.4pt}

\subsection{Locality Injection Matters}\label{sect-local-inject}
\setlength{\tabcolsep}{4pt}
\begin{table*}
	\caption{Studies on the effect of Locality Injection. The throughput is tested on the same 2080Ti GPU with a batch size of 128 and measured in samples/second. Note the throughput is observably reduced by adding the conv layers with negligible FLOPs.}
	\label{table-local-inject-matter}
	\vspace{-0.25in}
	\begin{center}
		\small
		\begin{tabular}{l|lcccccccc}
			\hline
			Dataset &Model			&	$1\times1$ conv	&	$3\times3$ conv	&	Top-1 acc&	FLOPs 	&	Params 	&	Throughput\\
			\hline
			\multirow{4}{45pt}{CIFAR-100}	&	ResMLP-12		&					&					&	55.58	&	1812M	&	7.1M	&	2561	\\
			&ResMLP-12		&	\checkmark		&					&	57.96	&	+786K	&	+60		&	2093	\\	% (1+4) * 12
			&ResMLP-12		&					&	\checkmark		&	63.76	&	+7077K	&	+156	&	1858	\\	% (9+4) * 12
			&ResMLP-12		&	\checkmark		&	\checkmark		&	65.09	&	+7864K	&	+216	&	1619	\\			
			\hline
			\multirow{4}{45pt}{CIFAR-100} &RepMLPNet		&					&					&	59.07	&	468M	&	8.3M	&	6273\\
			&RepMLPNet		&	\checkmark		&					&	60.73	&	+294K	&	+720	&	5872\\
			&RepMLPNet		&					&	\checkmark		&	65.36	&	+2654K	&	+2640	&	5721\\				
			&RepMLPNet		&	\checkmark		&	\checkmark		&	67.43	&	+2949K	&	+3360	&	5328\\
			\hline
			\multirow{4}{45pt}{ImageNet}&	RepMLPNet-T224	&	&	&	74.33	&	2.79B	&	38.3M	&	1709	\\
			&RepMLPNet-T224	&	\checkmark	&\checkmark				&	76.48	&	+10M	&	+20K	&	1354	\\
			&RepMLPNet-D256	&				&						&	78.58	&	8.61B	&	86.94M	&	715		\\			
			&RepMLPNet-D256	&	\checkmark	&\checkmark				&	80.88	&	+26M	&	+60k	&	570	\\
			\hline
		\end{tabular}
	\end{center}
	\vspace{-0.35in}
\end{table*}
\setlength{\tabcolsep}{1.4pt}

We conduct ablation studies on CIFAR-100~\cite{krizhevsky2009learning} and ImageNet to evaluate Locality Injection. Specifically, we build a small RepMLPNet with two stages, $\vect{B}=[6,6]$, $\vect{S}=[8,32]$, $C=128$. As another benchmark model, we scale down ResMLP-12 by reducing the channel dimensions. Besides, we change the downsampling ratio at the very beginning of all the models to 2$\times$, so that the embedding dimension becomes 16$\times$16, which is close to the case of ResMLP designed for ImageNet (14$\times$14). %All the structural hyper-parameters are casually set since we do not intend to chase the state-of-the-art on such a small dataset. 

For the ResMLP, we add 1$\times$1 and 3$\times$3 branches parallel to the spatial aggregation layer. Note that the spatial aggregation in ResMLP and MLP-Mixer is equivalent to our set-sharing FC with only one share-set (\ie, all the channels use the same set of $(hw)^2$ parameters). In this case, all the conv layers should have $s=1$ accordingly, which means a depth-wise conv with one input channel and one output channel. Consequently, adding a 1$\times$1 conv introduces only \emph{five parameters} (one for the 1$\times$1$\times$1$\times$1 kernel and four for the single-channel BN layer including $\mu,\sigma,\gamma,\beta$), so the whole model has only $5\times12=60$ extra parameters. Similarly, adding a 3$\times$3 layer brings only $(3\times3+4)\times12=156$ parameters. Though the extra parameters and FLOPs are negligible, the speed is observably slowed down (\eg, with 1$\times$1 and 3$\times$3 conv, the \emph{training-time} ResMLP-12 has only 0.4\% higher FLOPs but runs 37\% slower) due to the reduction of degree of parallelism~\cite{ma2018shufflenet}, which highlights the significance of merging the conv layers into the FC. For the RepMLPNet, the extra parameters brought by adding a 3$\times$3 conv is $(3\times3+4)s$ due to the set-sharing mechanism. 

All the ResMLPs and RepMLPNets on CIFAR-100 are trained with the same learning rate schedule and weight decay as described before, a batch size of 128 on a single GPU, and the standard data augmentation: padding to 40$\times$40, randomly cropping to 32$\times$32 and left-right random flipping. Predictably, the performance is not satisfactory since CIFAR-100 is too small for the FC layer to learn the inductive bias from the data. This discovery is consistent with the concurrent works~\cite{tolstikhin2021mlp,touvron2021resmlp} which highlight that MLPs show inferior performance on small datasets. Adding the conv branches only during training significantly boost the accuracy even though they are eventually eliminated. Impressively, though the ResMLP has only \emph{216 extra training-time parameters}, \emph{the accuracy raises by 9.51\%}, and we observe a similar phenomenon on RepMLPNet. We then experiment with RepMLP-T224/D256 on ImageNet. With the Local Perceptron removed, the accuracy decreases by 2.15\% and 2.30\%, respectively. The gap is narrower compared to the results on CIFAR, which is expected because ImageNet is significantly larger, allowing the model to learn some inductive bias from data \cite{tolstikhin2021mlp}). In summary, as Locality Injection works on different models and datasets, we conclude that it is a universal tool for vision MLPs.

\subsection{Semantic Segmentation}

The hierarchical design of RepMLPNet qualifies it as a backbone for the off-the-shelf downstream frameworks that require feature maps of different levels, \eg, UperNet~\cite{xiao2018unified}. However, transferring an ImageNet-pretrained MLP to the downstream task is challenging. Taking Cityscapes~\cite{cityscapes} semantic segmentation as the example, we reckon there are two primary obstacles for using an MLP backbone. \textbf{1)} The backbone is usually trained with low resolution (\eg, 256$\times$256 on ImageNet) then transferred to the high-resolution task (\eg, 1024$\times$2048 of Cityscapes). However, the parameter count of MLP is coupled with the input resolution (by our specific definition of ``MLP''), making the transfer difficult. \textbf{2)} The resolution for training (512$\times$1024 on Cityscapes) and testing (1024$\times$2048) may not be the same, so the backbone has to adapt to a variable resolution. 

In brief, our solution is to split the inputs into non-overlapping patches, feed the patches into the backbone, restore the output patches to form the feature maps, which are then fed into the downstream frameworks. For example, the first RepMLP Block of RepMLPNet-D256 works with 64$\times$64 inputs because the FC kernel is (64$^2$, 64$^2$). On Cityscapes, we can split the feature map into several 64$\times$64 patches and feed the patches into the RepMLP Block. However, doing so breaks the correlations among patches hence hinders a global understanding of the semantic information. Accordingly, we propose to replace the embedding (2$\times$2 conv) layers by regular conv (3$\times$3 conv) layers to communicate information across the patch borders. Interestingly, one may worry such a strategy would yield poor results at the edges of patches, but we observe that \emph{the predictions at the edges are as good as the other parts} (see the Appendix).

As the first attempt to use an MLP as the backbone for Cityscapes semantic segmentation, RepMLPNet delivers promising results (Table~\ref{table-seg}). By further replacing the 3$\times$3 downsampling layers by 5$\times$5, the mIoU improves with negligible extra FLOPs, which is expected as a larger convolution enables better communications among patches. Specifically, we use the implementation of UperNet~\cite{xiao2018unified} in MMSegmentation \cite{mmseg2020} and the 80K-iteration training schedule. We present the details and analysis in the Appendix.

\setlength{\tabcolsep}{4pt}
\begin{table}
	\caption{Results on Cityscapes val set. By replacing the 3$\times$3 downsampling layers with 5$\times$5, the mIoU further increases. }
	\label{table-seg}
	\vspace{-0.25in}
	\begin{center}
		\small
		\begin{tabular}{lccccccccc}
			\hline
			Backbone				&	FLOPs			&	mIoU		\\
			\hline
			ResNet-101				&	2049.82G		&	79.03		\\
			RepMLPNet-D256		 	&	1960.01G		&	76.27		\\
			RepMLPNet-D256 (conv5) 	&	1960.16G		&	77.12		\\
			\hline
		\end{tabular}
	\end{center}
\vspace{-0.3in}
\end{table}
\setlength{\tabcolsep}{1.4pt}

\section{Limitations and Conclusions}

This paper proposes a re-parameterization method to inject locality into FC layers, a novel MLP-style block, and a hierarchical MLP architecture. RepMLPNet is favorable compared to several concurrently proposed MLPs in terms of the accuracy-efficiency trade-off and the training costs. 

However, as an MLP, RepMLPNet has several noticeable common weaknesses. \textbf{1)} Similar to the Vision Transformers, MLPs are easy to overfit, requiring strong data augmentation and regularization techniques. \textbf{2)} On the low-power devices like mobile phones, the model size of MLPs may be an obstacle. \textbf{3)} Though the results of our first attempt to use MLP backbone for semantic segmentation are promising, we observe no superiority over the traditional CNNs.

%%%%%%%%% REFERENCES
{\small
\bibliographystyle{ieee_fullname}
\bibliography{repmlpbib}

\begin{thebibliography}{10}\itemsep=-1pt

\bibitem{cao2019gcnet}
Yue Cao, Jiarui Xu, Stephen Lin, Fangyun Wei, and Han Hu.
\newblock Gcnet: Non-local networks meet squeeze-excitation networks and
  beyond.
\newblock In {\em Proceedings of the IEEE/CVF International Conference on
  Computer Vision Workshops}, pages 0--0, 2019.

\bibitem{im2col}
Kumar Chellapilla, Sidd Puri, and Patrice Simard.
\newblock High performance convolutional neural networks for document
  processing.
\newblock In {\em Tenth International Workshop on Frontiers in Handwriting
  Recognition}. Suvisoft, 2006.

\bibitem{chen2021cyclemlp}
Shoufa Chen, Enze Xie, Chongjian Ge, Ding Liang, and Ping Luo.
\newblock Cyclemlp: A mlp-like architecture for dense prediction.
\newblock {\em arXiv preprint arXiv:2107.10224}, 2021.

\bibitem{cho2017mec}
Minsik Cho and Daniel Brand.
\newblock Mec: memory-efficient convolution for deep neural network.
\newblock In {\em International Conference on Machine Learning}, pages
  815--824. PMLR, 2017.

\bibitem{mmseg2020}
MMSegmentation Contributors.
\newblock {MMSegmentation}: Openmmlab semantic segmentation toolbox and
  benchmark.
\newblock \url{https://github.com/open-mmlab/mmsegmentation}, 2020.

\bibitem{cityscapes}
Marius Cordts, Mohamed Omran, Sebastian Ramos, Timo Rehfeld, Markus Enzweiler,
  Rodrigo Benenson, Uwe Franke, Stefan Roth, and Bernt Schiele.
\newblock The cityscapes dataset for semantic urban scene understanding.
\newblock In {\em 2016 {IEEE} Conference on Computer Vision and Pattern
  Recognition, {CVPR} 2016, Las Vegas, NV, USA, June 27-30, 2016}, pages
  3213--3223. {IEEE} Computer Society, 2016.

\bibitem{cubuk2020randaugment}
Ekin~D Cubuk, Barret Zoph, Jonathon Shlens, and Quoc~V Le.
\newblock Randaugment: Practical automated data augmentation with a reduced
  search space.
\newblock In {\em Proceedings of the IEEE/CVF Conference on Computer Vision and
  Pattern Recognition Workshops}, pages 702--703, 2020.

\bibitem{deng2009imagenet}
Jia Deng, Wei Dong, Richard Socher, Li-Jia Li, Kai Li, and Li Fei-Fei.
\newblock Imagenet: A large-scale hierarchical image database.
\newblock In {\em Computer Vision and Pattern Recognition, 2009. CVPR 2009.
  IEEE Conference on}, pages 248--255. IEEE, 2009.

\bibitem{ding2019centripetal}
Xiaohan Ding, Guiguang Ding, Yuchen Guo, and Jungong Han.
\newblock Centripetal sgd for pruning very deep convolutional networks with
  complicated structure.
\newblock In {\em Proceedings of the IEEE Conference on Computer Vision and
  Pattern Recognition}, pages 4943--4953, 2019.

\bibitem{ding2019approximated}
Xiaohan Ding, Guiguang Ding, Yuchen Guo, Jungong Han, and Chenggang Yan.
\newblock Approximated oracle filter pruning for destructive cnn width
  optimization.
\newblock In {\em International Conference on Machine Learning}, pages
  1607--1616, 2019.

\bibitem{ding2019acnet}
Xiaohan Ding, Yuchen Guo, Guiguang Ding, and Jungong Han.
\newblock Acnet: Strengthening the kernel skeletons for powerful cnn via
  asymmetric convolution blocks.
\newblock In {\em Proceedings of the IEEE International Conference on Computer
  Vision}, pages 1911--1920, 2019.

\bibitem{ding2021manipulating}
Xiaohan Ding, Tianxiang Hao, Jungong Han, Yuchen Guo, and Guiguang Ding.
\newblock Manipulating identical filter redundancy for efficient pruning on
  deep and complicated cnn.
\newblock {\em arXiv preprint arXiv:2107.14444}, 2021.

\bibitem{ding2021resrep}
Xiaohan Ding, Tianxiang Hao, Jianchao Tan, Ji Liu, Jungong Han, Yuchen Guo, and
  Guiguang Ding.
\newblock Resrep: Lossless cnn pruning via decoupling remembering and
  forgetting.
\newblock In {\em Proceedings of the IEEE/CVF International Conference on
  Computer Vision}, pages 4510--4520, 2021.

\bibitem{ding2021diverse}
Xiaohan Ding, Xiangyu Zhang, Jungong Han, and Guiguang Ding.
\newblock Diverse branch block: Building a convolution as an inception-like
  unit.
\newblock In {\em Proceedings of the IEEE/CVF Conference on Computer Vision and
  Pattern Recognition}, pages 10886--10895, 2021.

\bibitem{ding2021repvgg}
Xiaohan Ding, Xiangyu Zhang, Ningning Ma, Jungong Han, Guiguang Ding, and Jian
  Sun.
\newblock Repvgg: Making vgg-style convnets great again.
\newblock {\em arXiv preprint arXiv:2101.03697}, 2021.

\bibitem{ding2022scaling}
Xiaohan Ding, Xiangyu Zhang, Yizhuang Zhou, Jungong Han, Guiguang Ding, and
  Jian Sun.
\newblock Scaling up your kernels to 31x31: Revisiting large kernel design in
  cnns.
\newblock {\em arXiv preprint arXiv:2203.06717}, 2022.

\bibitem{ding2019global}
Xiaohan Ding, Xiangxin Zhou, Yuchen Guo, Jungong Han, Ji Liu, et~al.
\newblock Global sparse momentum sgd for pruning very deep neural networks.
\newblock {\em Advances in Neural Information Processing Systems}, 32, 2019.

\bibitem{dosovitskiy2020image}
Alexey Dosovitskiy, Lucas Beyer, Alexander Kolesnikov, Dirk Weissenborn,
  Xiaohua Zhai, Thomas Unterthiner, Mostafa Dehghani, Matthias Minderer, Georg
  Heigold, Sylvain Gelly, et~al.
\newblock An image is worth 16x16 words: Transformers for image recognition at
  scale.
\newblock {\em arXiv preprint arXiv:2010.11929}, 2020.

\bibitem{han2015learning}
Song Han, Jeff Pool, John Tran, and William Dally.
\newblock Learning both weights and connections for efficient neural network.
\newblock In {\em Advances in Neural Information Processing Systems}, pages
  1135--1143, 2015.

\bibitem{he2016deep}
Kaiming He, Xiangyu Zhang, Shaoqing Ren, and Jian Sun.
\newblock Deep residual learning for image recognition.
\newblock In {\em Proceedings of the IEEE conference on computer vision and
  pattern recognition}, pages 770--778, 2016.

\bibitem{hendrycks2016gaussian}
Dan Hendrycks and Kevin Gimpel.
\newblock Gaussian error linear units (gelus).
\newblock {\em arXiv preprint arXiv:1606.08415}, 2016.

\bibitem{ioffe2015batch}
Sergey Ioffe and Christian Szegedy.
\newblock Batch normalization: Accelerating deep network training by reducing
  internal covariate shift.
\newblock In {\em International Conference on Machine Learning}, pages
  448--456, 2015.

\bibitem{krizhevsky2009learning}
Alex Krizhevsky and Geoffrey Hinton.
\newblock Learning multiple layers of features from tiny images.
\newblock 2009.

\bibitem{winograd}
Andrew Lavin and Scott Gray.
\newblock Fast algorithms for convolutional neural networks.
\newblock In {\em Proceedings of the IEEE Conference on Computer Vision and
  Pattern Recognition}, pages 4013--4021, 2016.

\bibitem{li2016pruning}
Hao Li, Asim Kadav, Igor Durdanovic, Hanan Samet, and Hans~Peter Graf.
\newblock Pruning filters for efficient convnets.
\newblock {\em arXiv preprint arXiv:1608.08710}, 2016.

\bibitem{liu2021pay}
Hanxiao Liu, Zihang Dai, David~R So, and Quoc~V Le.
\newblock Pay attention to mlps.
\newblock {\em arXiv preprint arXiv:2105.08050}, 2021.

\bibitem{liu2021swin}
Ze Liu, Yutong Lin, Yue Cao, Han Hu, Yixuan Wei, Zheng Zhang, Stephen Lin, and
  Baining Guo.
\newblock Swin transformer: Hierarchical vision transformer using shifted
  windows.
\newblock In {\em Proceedings of the IEEE/CVF International Conference on
  Computer Vision}, pages 10012--10022, 2021.

\bibitem{liu2019rethinking}
Zhuang Liu, Mingjie Sun, Tinghui Zhou, Gao Huang, and Trevor Darrell.
\newblock Rethinking the value of network pruning.
\newblock In Yoshua Bengio and Yann LeCun, editors, {\em 7th International
  Conference on Learning Representations, {ICLR} 2019, New Orleans, LA, USA,
  May 6-9, 2019}. OpenReview.net, 2019.

\bibitem{loshchilov2017decoupled}
Ilya Loshchilov and Frank Hutter.
\newblock Decoupled weight decay regularization.
\newblock {\em arXiv preprint arXiv:1711.05101}, 2017.

\bibitem{ma2018shufflenet}
Ningning Ma, Xiangyu Zhang, Hai-Tao Zheng, and Jian Sun.
\newblock Shufflenet v2: Practical guidelines for efficient cnn architecture
  design.
\newblock In {\em Proceedings of the European conference on computer vision
  (ECCV)}, pages 116--131, 2018.

\bibitem{fft-conv}
Michael Mathieu, Mikael Henaff, and Yann LeCun.
\newblock Fast training of convolutional networks through ffts.
\newblock {\em arXiv preprint arXiv:1312.5851}, 2013.

\bibitem{paszke2019pytorch}
Adam Paszke, Sam Gross, Francisco Massa, Adam Lerer, James Bradbury, Gregory
  Chanan, Trevor Killeen, Zeming Lin, Natalia Gimelshein, Luca Antiga, et~al.
\newblock Pytorch: An imperative style, high-performance deep learning library.
\newblock {\em arXiv preprint arXiv:1912.01703}, 2019.

\bibitem{regnet}
Ilija Radosavovic, Raj~Prateek Kosaraju, Ross Girshick, Kaiming He, and Piotr
  Doll{\'a}r.
\newblock Designing network design spaces.
\newblock In {\em Proceedings of the IEEE/CVF Conference on Computer Vision and
  Pattern Recognition}, pages 10428--10436, 2020.

\bibitem{simonyan2014very}
Karen Simonyan and Andrew Zisserman.
\newblock Very deep convolutional networks for large-scale image recognition.
\newblock {\em arXiv preprint arXiv:1409.1556}, 2014.

\bibitem{efficientnet}
Mingxing Tan and Quoc~V Le.
\newblock Efficientnet: Rethinking model scaling for convolutional neural
  networks.
\newblock {\em arXiv preprint arXiv:1905.11946}, 2019.

\bibitem{tolstikhin2021mlp}
Ilya Tolstikhin, Neil Houlsby, Alexander Kolesnikov, Lucas Beyer, Xiaohua Zhai,
  Thomas Unterthiner, Jessica Yung, Daniel Keysers, Jakob Uszkoreit, Mario
  Lucic, et~al.
\newblock Mlp-mixer: An all-mlp architecture for vision.
\newblock {\em arXiv preprint arXiv:2105.01601}, 2021.

\bibitem{touvron2021resmlp}
Hugo Touvron, Piotr Bojanowski, Mathilde Caron, Matthieu Cord, Alaaeldin
  El-Nouby, Edouard Grave, Gautier Izacard, Armand Joulin, Gabriel Synnaeve,
  Jakob Verbeek, et~al.
\newblock Resmlp: Feedforward networks for image classification with
  data-efficient training.
\newblock {\em arXiv preprint arXiv:2105.03404}, 2021.

\bibitem{touvron2021training}
Hugo Touvron, Matthieu Cord, Matthijs Douze, Francisco Massa, Alexandre
  Sablayrolles, and Herv{\'e} J{\'e}gou.
\newblock Training data-efficient image transformers \& distillation through
  attention.
\newblock In {\em International Conference on Machine Learning}, pages
  10347--10357. PMLR, 2021.

\bibitem{vaswani2017attention}
Ashish Vaswani, Noam Shazeer, Niki Parmar, Jakob Uszkoreit, Llion Jones,
  Aidan~N Gomez, Lukasz Kaiser, and Illia Polosukhin.
\newblock Attention is all you need.
\newblock {\em arXiv preprint arXiv:1706.03762}, 2017.

\bibitem{wang2018non}
Xiaolong Wang, Ross Girshick, Abhinav Gupta, and Kaiming He.
\newblock Non-local neural networks.
\newblock In {\em Proceedings of the IEEE conference on computer vision and
  pattern recognition}, pages 7794--7803, 2018.

\bibitem{xiao2018unified}
Tete Xiao, Yingcheng Liu, Bolei Zhou, Yuning Jiang, and Jian Sun.
\newblock Unified perceptual parsing for scene understanding.
\newblock In {\em Proceedings of the European Conference on Computer Vision
  (ECCV)}, pages 418--434, 2018.

\bibitem{xie2017aggregated}
Saining Xie, Ross Girshick, Piotr Doll{\'a}r, Zhuowen Tu, and Kaiming He.
\newblock Aggregated residual transformations for deep neural networks.
\newblock In {\em Proceedings of the IEEE conference on computer vision and
  pattern recognition}, pages 1492--1500, 2017.

\bibitem{yu2021s}
Tan Yu, Xu Li, Yunfeng Cai, Mingming Sun, and Ping Li.
\newblock S2-mlp: Spatial-shift mlp architecture for vision.
\newblock {\em arXiv preprint arXiv:2106.07477}, 2021.

\bibitem{yun2019cutmix}
Sangdoo Yun, Dongyoon Han, Seong~Joon Oh, Sanghyuk Chun, Junsuk Choe, and
  Youngjoon Yoo.
\newblock Cutmix: Regularization strategy to train strong classifiers with
  localizable features.
\newblock In {\em Proceedings of the IEEE/CVF International Conference on
  Computer Vision}, pages 6023--6032, 2019.

\bibitem{zhang2017mixup}
Hongyi Zhang, Moustapha Cisse, Yann~N Dauphin, and David Lopez-Paz.
\newblock mixup: Beyond empirical risk minimization.
\newblock {\em arXiv preprint arXiv:1710.09412}, 2017.

\end{thebibliography}
}

\newpage

\section*{Appendix A: Visualizing Locality Injection}

We demonstrate the locality is injected into the FC kernel by showing the kernel weights. 

Specifically, we visualize the weights of FC3 kernel sampled from the 10th RepMLP Block of the 3rd stage of RepMLPNet-D256 trained on ImageNet. We reshape the kernel into $\bar{\mathrm{W}}(s,h,w,1,h,w)$, which is $(16, 16, 16, 1, 16, 16)$, then sample the weights related to the first input channel and the (7,7) point (marked by a purple square) on the first output channel, which is $\bar{\mathrm{W}}_{1,7,7,1,:,:}$ (indices starting from 1). Then we take the absolute value, rescale by the minimum value of the whole matrix, and take the logarithm for the better readability. 

\begin{figure}
	\begin{center}
		\includegraphics[width=\linewidth]{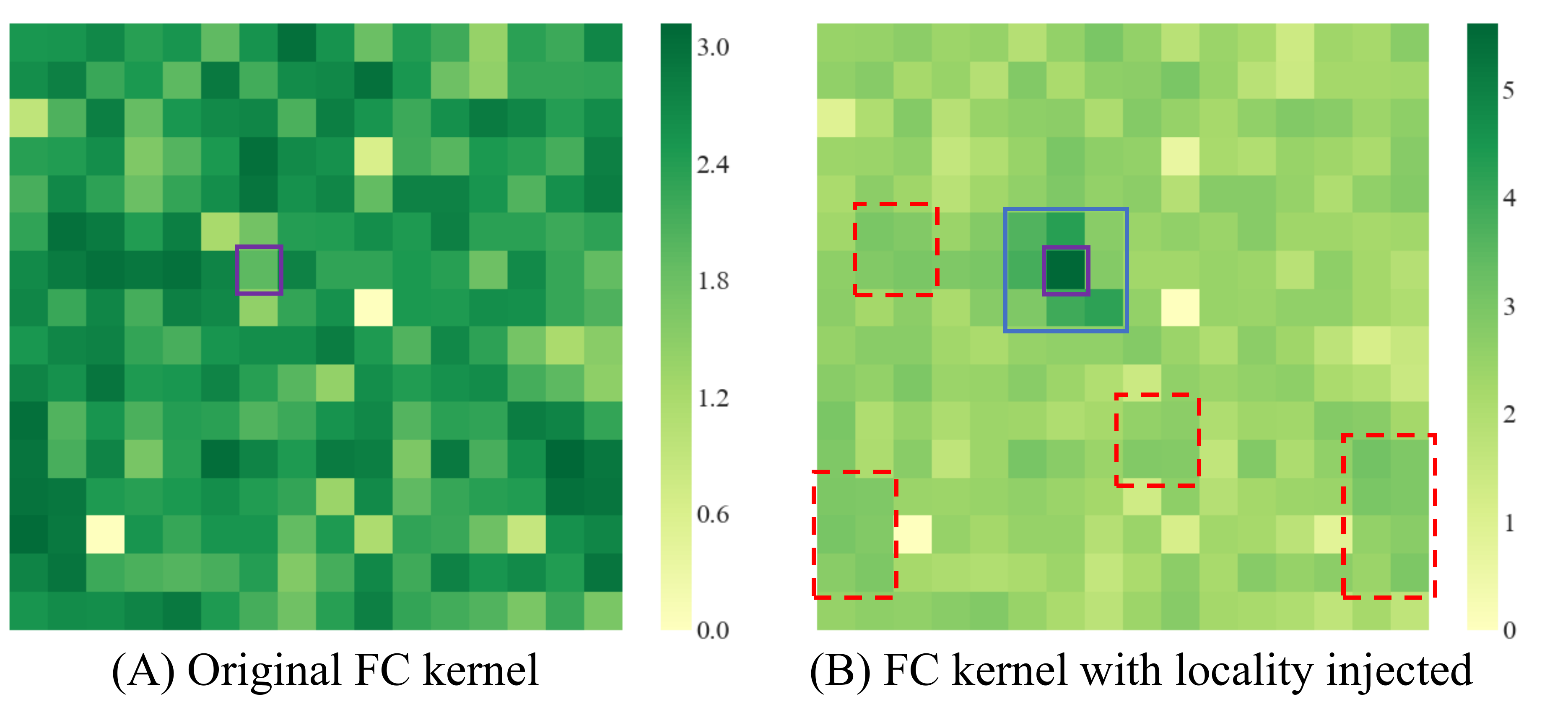}
		\vspace{-0.25in}
		\caption{FC kernel before and after Locality Injection.}
		\label{fig-visualize}
		\vspace{-0.25in}
	\end{center}
\end{figure}

As shown in Fig.~\ref{fig-visualize}, a point with darker color indicates the FC considers the corresponding position on the input channel more related to the output point at (7,7). Obviously, the original kernel has no locality as the marked point and the neighbors have no larger values than the others. 

Then we merge the parallel conv layers into the kernel via Locality Injection. The resultant kernel has larger values around the marked point, suggesting that the model focuses more on the neighbors, which is expected. Besides, the kernel's capacity to model the long-range dependencies is not lost as some points (marked by red dashed rectangles) outside the largest conv kernel (3$\times$3 in this case, marked by a blue square) still have larger values than some points inside.

\section*{Appendix B: Details of Semantic Segmentation}

We solve the problem of using MLP as the backbone for semantic segmentation (\eg, UperNet) by \textbf{1)} the hierarchical design, \textbf{2)} splitting feature maps into non-overlapping patches and \textbf{3)} communications between patches.

\begin{figure}
	\begin{center}
		\includegraphics[width=\linewidth]{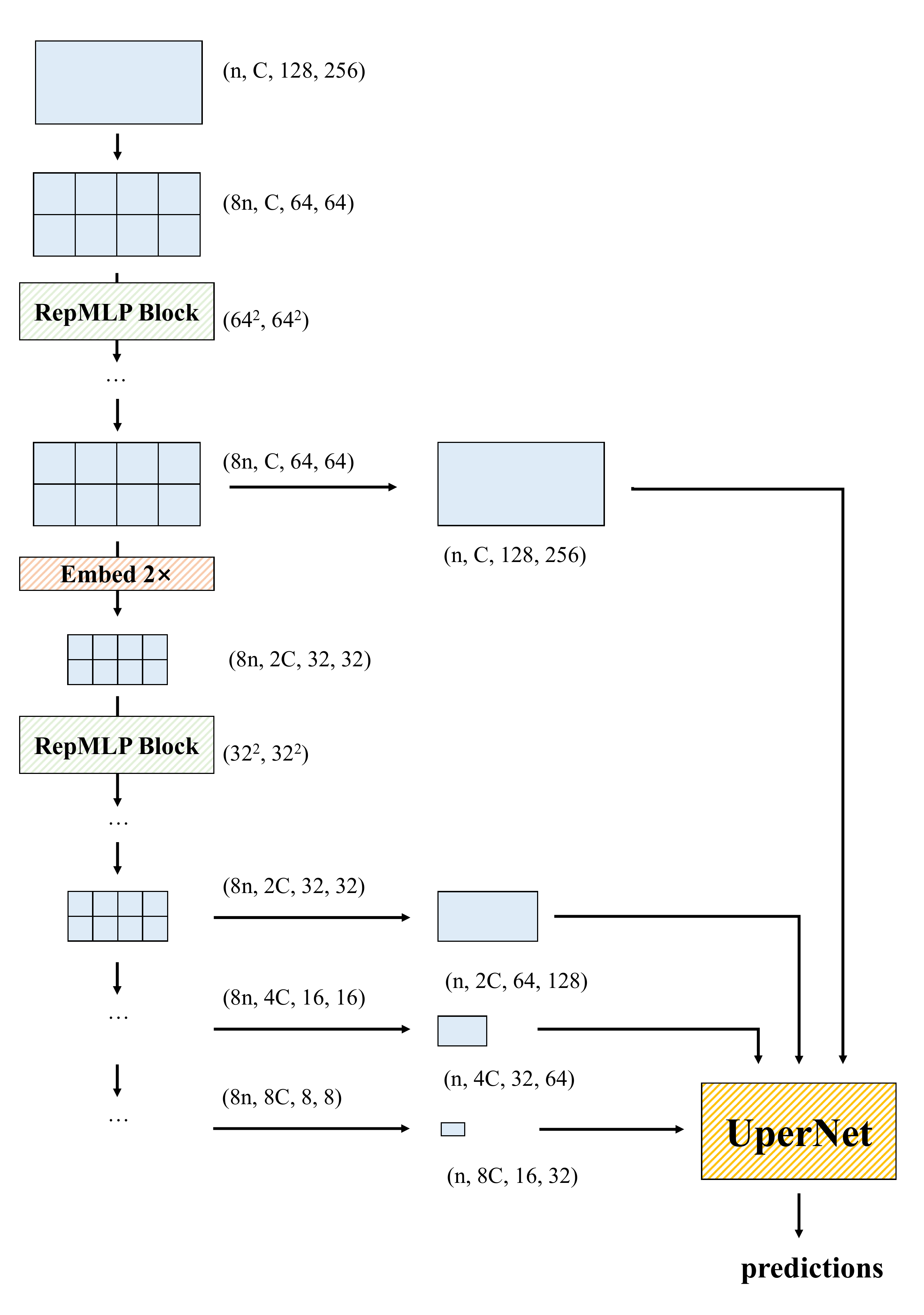}
		\vspace{-0.25in}
		\caption{An example of using RepMLPNet as the backbone of UperNet. After 4$\times$ downsampling on the 512$\times$1024 inputs, the feature map size becomes 128$\times$256. Then we split the feature maps into 2$\times$4 non-overlapping patches, each of 64$\times$64, because the first RepMLP Block maps inputs of 64$\times$64 into 64$\times$64 (\ie, the FC kernel is (64$^2$,64$^2$)). Similarly, the outputs of the four stages are reshaped back and fed into the UperNet. }
		\label{fig-sketch}
		\vspace{-0.25in}
	\end{center}
\end{figure}

Fig. \ref{fig-sketch} shows an example of RepMLPNet + UperNet. 

\begin{figure}
	\begin{center}
		\includegraphics[width=\linewidth]{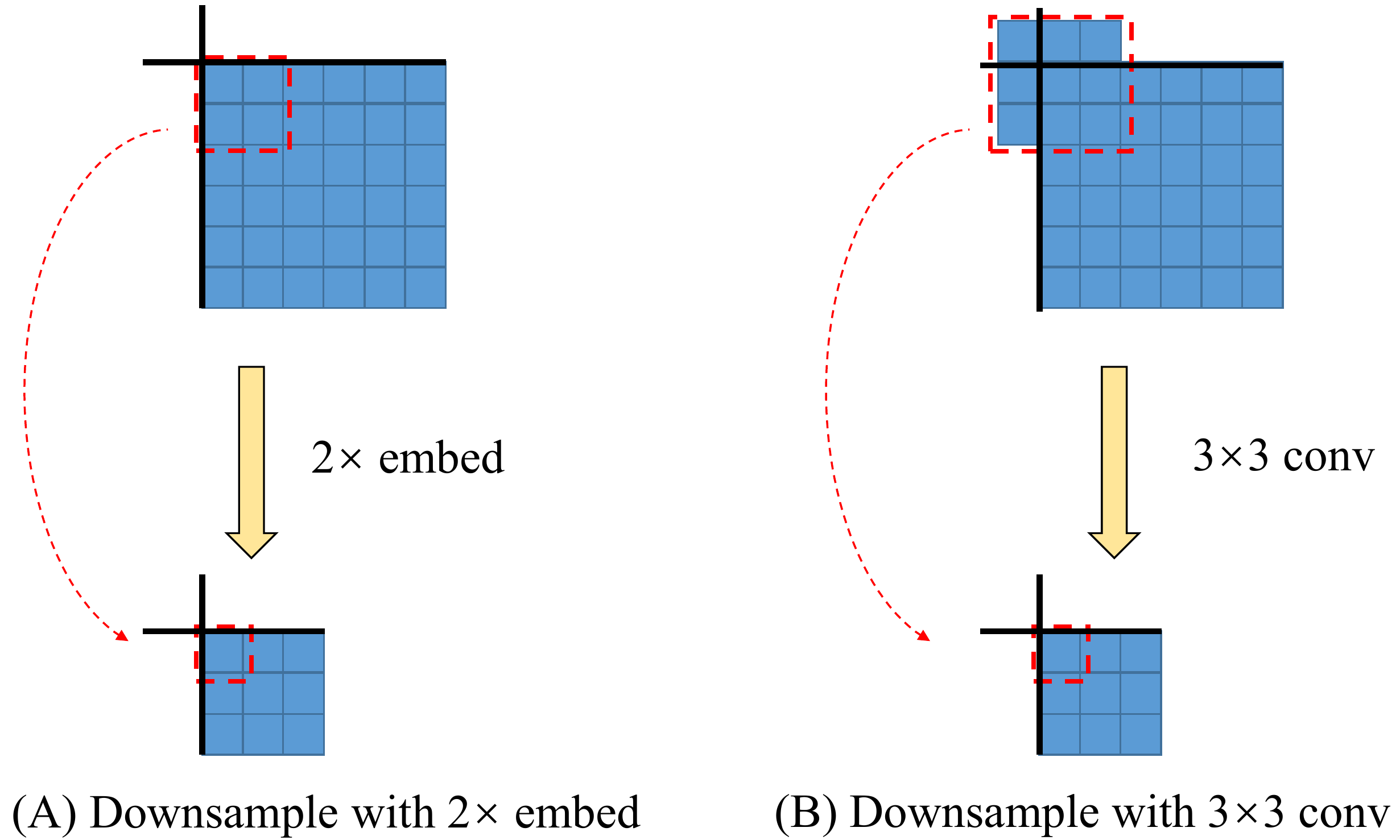}
		\vspace{-0.25in}
		\caption{The difference between 2$\times$ embedding and 3$\times$3 conv is that the latter realizes communications across patch edges. In this figure, a square denotes a pixel on a feature map and the thick lines denote the edges of patches. We take the upper left corner of a patch as an example}
		\label{fig-downsample}
		\vspace{-0.25in}
	\end{center}
\end{figure}

\textbf{1)} UperNet requires feauture maps of 4 different levels, which fits our hierarchical architecture. \textbf{2)} Since RepMLP Block works with a fixed input feature map size, we split the feature maps into non-overlapping patches each with the required size. \textbf{3)} The original 2$\times$ embedding cannot realize inter-patch communication, so we replace it with 3$\times$3 conv. To reduce the computational cost of 3$\times$3 conv, we decompose it into a 1$\times$1 conv for expanding the channels and a 3$\times$3 stride-2 depth-wise conv for downsampling. Fig.~\ref{fig-downsample} shows the difference between a 2$\times$ embedding and a 3$\times$3 conv.

\begin{figure*}[t!]
	\centering
	\begin{subfigure}[t]{0.95\textwidth}
		\includegraphics[page=1,width=\textwidth]{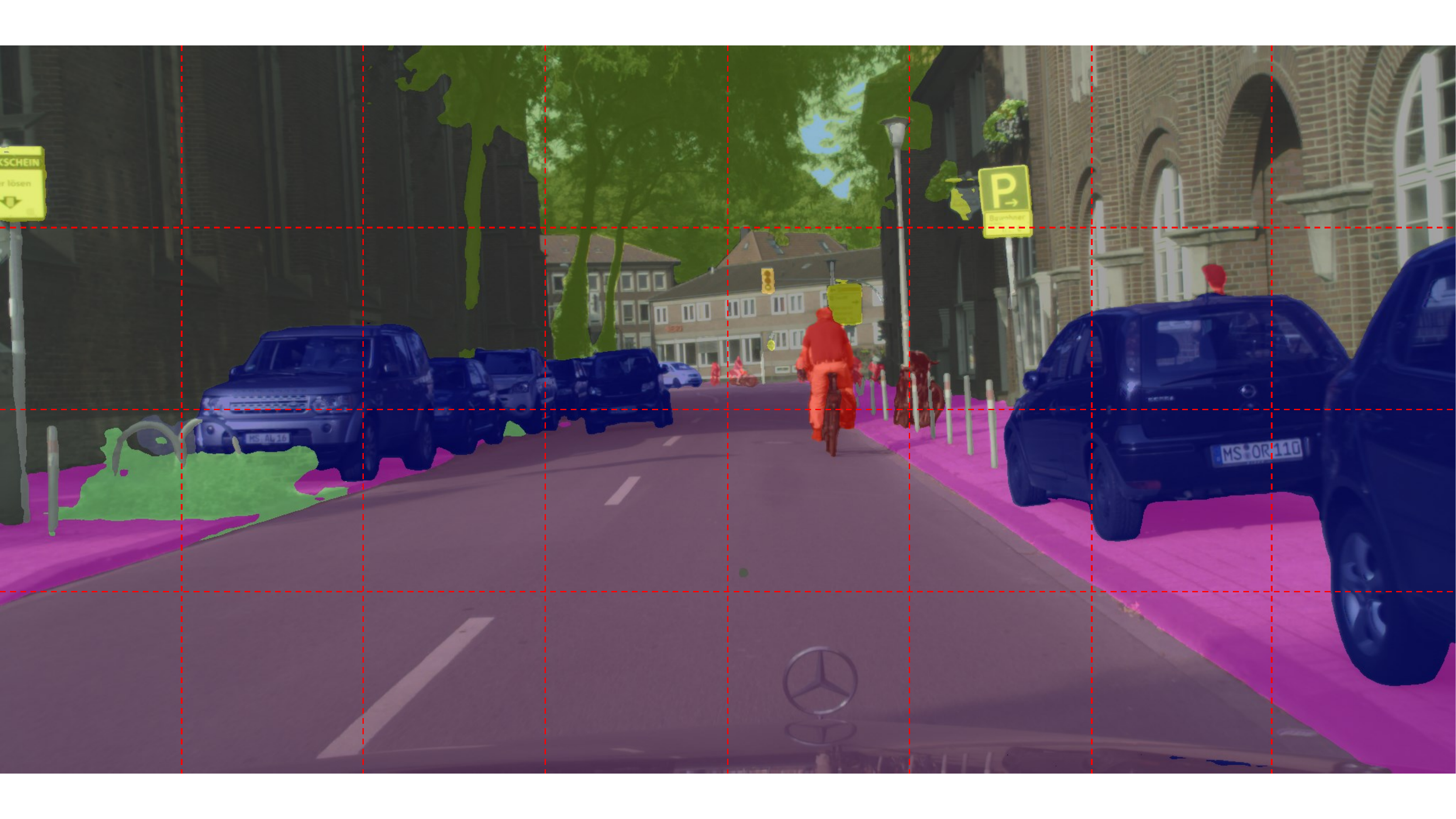}
		\label{fig:c}
	\end{subfigure}
	\begin{subfigure}[t]{0.95\textwidth}
		\includegraphics[page=2,width=\textwidth]{seg-result.pdf}
		\label{fig:f}
	\end{subfigure}
	\caption{The predictions at the edges of patches are no observably worse. We show two images from the Cityscapes validation set as examples. As the test resolution is 1024$\times$2048, the input to the first RepMLP Block is 256$\times$512 (after the beginning 4$\times$ downsampling), so that the input is split into 32 non-overlapping patches and then fed into RepMLP Blocks. We use red dashed lines to denote the edges of patches and it is observed that the predictions at the edges are almost as good as the other parts.}
	\label{fig-seg-result}
	\vskip -0.2in
\end{figure*}

Interestingly, as the input is split into non-overlapping patches, one may expect that the predictions would be less accurate at the edges of patches, but we observe no such phenomenon. Fig. \ref{fig-seg-result} shows that the predictions at the edges are as good as the internal pixels within patches. This discovery suggests that the dependencies across patches have been well established and that the representational capacity of MLP is strong enough for such a dense prediction task.

We hope our results spark further research on the application of MLP on downstream tasks.

\end{document}